\documentclass{article}

\usepackage{microtype}
\usepackage{graphicx}
\usepackage{subcaption}
\usepackage{booktabs}
\usepackage{hyperref}
\usepackage[textsize=tiny]{todonotes}

\usepackage[accepted]{icml2026}

\usepackage{amsmath}
\usepackage{amssymb}
\usepackage{mathtools}
\usepackage{amsthm}

\usepackage[capitalize,noabbrev]{cleveref}

\icmltitlerunning{Whose Alignment? Process Alignment Across Organizational Contexts}

\begin{document}

\twocolumn[
  \icmltitle{Whose Alignment? Comparing LLM Process Alignment\\
    Across Diverse Organizational Decision Contexts}

  \begin{icmlauthorlist}
    \icmlauthor{Niklas Weller}{yyy}
    \icmlauthor{Emilio Barkett}{comp}
  \end{icmlauthorlist}

  \icmlaffiliation{yyy}{University of St. Gallen, Switzerland}
  \icmlaffiliation{comp}{Columbia University, New York, NY USA}

  \icmlcorrespondingauthor{Niklas Weller}{niklas.weller@unisg.ch}
  \icmlcorrespondingauthor{Emilio Barkett}{eab2291@columbia.edu}

  \icmlkeywords{pluralistic alignment, LLM alignment, LLM decision-making, cue utilization, organizational alignment}

  \vskip 0.3in
]

\printAffiliationsAndNotice{\icmlEqualContribution}


\begin{abstract}
Steerable pluralism requires a model to faithfully represent one specified perspective. Organizations are a natural setting for this demand, since they deploy LLMs to make decisions that must reflect their own policy. Yet, most existing work fixes that perspective at the level of individuals or demographic groups. We rely on a decision-policy capturing method to measure \emph{process alignment} in organizational settings, assessing whether an LLM faithfully reproduces the organization's decision policy rather than merely reaching the same conclusions. We find heterogeneity along two axes. Across models, baseline alignment varies strongly and tracks neither pricing nor general benchmark performance. Across organizations, the structure of alignment changes. In ECHR Article~6 decisions, process alignment predicts output accuracy ($r = 0.85$, $p < .001$), and making the organization's past decision policy explicit improves poorly aligned models. In consumer credit decisions, process alignment is low overall but varies more than output accuracy, and the models resist adopting the organization's weighting of protected attributes. Because historical credit decisions encode potentially discriminatory patterns, higher alignment there is not always desirable. Process-level measurement is therefore necessary, and depending on whether the target policy is normatively desirable, the same procedure can calibrate or audit a model. Deciding which policy to align to, and whether higher alignment is feasible or desirable, makes organizational alignment a pluralistic problem in its own right.
\end{abstract}


\section{Introduction}

Pluralistic alignment asks how AI systems can faithfully serve stakeholders with diverse, conflicting values rather than converging on a single dominant worldview \citep{sorensen2024roadmap}. Most work in this space focuses on \emph{between-person} plurality: aggregating individual preferences, handling annotation disagreements, or enabling models to adopt different user perspectives. We argue that \emph{between-organization} plurality is an equally important but largely neglected dimension. Organizations encode values as historically contingent decision policies, and these policies differ systematically across organizations, domains, and time. Artificial intelligence (AI) models vary in which organizational value systems they implicitly align with, and this variation is not captured by any existing pluralistic alignment evaluation framework. Recognizing between-organization plurality expands the scope of pluralistic alignment from \emph{whose individual preferences?} to \emph{whose institutional norms?} This question has direct consequences for how AI systems are deployed, audited, and governed in high-stakes settings.

Organizations are repositories and integrators of knowledge accumulated through experience, routines, culture, and institutional history \citep{walsh1991organizational,grant1996toward}. Some of this knowledge is explicit: rules, policies, codified procedures. Much is tacit: embedded in practice and professional judgment, resistant to full articulation \citep{polanyi1966tacit,pakarien2023relational, nonaka1994dynamic}. When organizations deploy large language models (LLMs) to decide in line with their intentions and values, they implicitly assume they can transfer this knowledge to a model. The tacit dimension resists explicit transfer.

This creates a problem that reaches beyond output accuracy. Due to differences in effective contextual knowledge, an LLM can reach the right decision for the wrong reasons, weighting information differently than the organization while still achieving adequate accuracy on the output metrics \citep{geirhos2020shortcut}. We refer to this divergence as process misalignment: a mismatch between organizational and LLM cue-utilization policies that output agreement alone may fail to detect.

To operationalize and measure decision process divergence, we rely on a methodology inspired by the Brunswik Lens Model \citep{brunswik1952conceptual,hammond1964psychology}. This allows us to estimate and compare the cue-weighting policy as a proxy for the decision process of both organizations and LLMs from observed decisions. We then compute cosine similarity between these policy vectors as the primary alignment metric. Similar procedures are well-established for studying expert judgment in clinical, educational, and managerial domains \citep{karelaia2008minds, karren2002review}. For example, \citet{smith2003factors} use its symmetric structure to compare normative guidelines for clinical depression-treatment against individual practitioners' prescribing policies, surfacing cue-weighting differences in the decision-policy. We adopt the same comparative move, with an organizational policy derived from past decisions in place of the guidelines (alignment target) and an LLM in place of the individual practitioner. In the following, we refer to this method as the \textit{Contextualized Alignment Lens Model} (CALM) and apply it in two domains:

\textbf{Study 1 (ECHR):} We analyze 1,000 ECHR Article~6 decisions from the collection released by \citet{chalkidis2021paragraph} across ten LLM models under three prompting conditions. We find sharp variation in baseline process alignment and show that explicit knowledge externalization substantially improves alignment for poorly-aligned models.

\textbf{Study 2 (German Credit):} We analyze decisions on consumer credit risk using the German Credit Dataset \citep{hofmann1994statlog}. Unlike the legal domain, the credit benchmark is derived from historical decisions that encode potentially discriminatory patterns. While the models perform poorly overall, we find that process alignment differs more than output accuracy and knowledge externalization fails to reliably improve alignment, which is partly caused by the models resisting to adjust their weighting of protected attributes. This is a distinctively pluralistic challenge: \emph{whose values} should the model align to?

Together, these studies reveal heterogeneity in organizational alignment along two axes. Across models, baseline alignment varies widely and tracks neither pricing nor general-purpose benchmark performance, so the choice of model is itself a choice among organizational value systems. Across organizations, the relationship between process alignment and output accuracy is unstable: strong and steerable in the legal domain, unreliable and resistant to intervention in the credit domain. The cross-domain contrast is the clearest instance of this between-organization plurality, the second axis our framework makes visible. This plurality forces a normative question that output-level evaluation cannot surface: which policy the model should align to, and whether higher alignment is desirable when the target encodes contested values. A related question, which we return to in the discussion, is whether a model's stated reasoning reflects the behavioral policy CALM estimates. We treat CALM as one instantiation of process-level measurement, and we expect that pluralistic alignment will require a family of such tools. We further argue that organizational steerability should become a design target for foundation models, and we outline a benchmark for it as future work.


\section{Background}

\subsection{The Brunswik Lens Model}

The operationalization of the Brunswik lens model \citep{brunswik1952conceptual,hammond1964psychology} represents judgment as a linear combination of observable cues: $J = \beta_0 + \sum_i \beta_i C_i$, where $C_i$ are cues and $\beta_i$ are learned weights. Originally developed for human decision-makers, it measures ecological validity (do cues genuinely predict outcomes?) and achievement (do the decision-maker's weights match ecological realities?). Applied to our setting of organizations and LLMs, it reveals whether they weight the same cues in the same direction and magnitude. This way, it provides a measure for whether their \emph{observed behavioral policies} are aligned. Meta-analytic evidence confirms that human judges systematically misweight cues relative to ecological validities \citep{karelaia2008minds}; the same divergence between stated and behavioral weights appears when comparing organizational guidelines to individual practitioners' actual decisions \citep{smith2003factors}.

\subsection{Pluralistic Alignment}
\citet{sorensen2024roadmap} distinguish three modes of pluralistic alignment: Overton pluralism (presenting a range of reasonable responses), distributional pluralism (reflecting population-level value distributions), and steerable pluralism (faithfully adopting a specified perspective when directed). Our proposed between-organization plurality maps most directly onto the steerable mode. Steerable pluralism emphasizes \emph{faithfulness}. In our case, this means that the model genuinely adopts the target policy rather than mimicking its surface outputs. We measure faithfulness behaviorally by testing whether a model's cue-weighting policy shifts toward the target organization's policy when making contextual knowledge explicit (following the terminology by \citet{nonaka1994dynamic} we refer to this as \textit{externalization}). The dominant approach to alignment via reinforcement learning from human feedback \citep{ouyang2022training} is often operationalized through aggregated human preferences, and therefore may encode a consensus value system rather than support institutional plurality. This reflects a broader challenge in AI alignment, where the target values are plural and contested rather than singular \citep{gabriel2020artificial}.

Prior work has shown that LLMs can achieve high accuracy on judgment tasks while relying on different behavioral strategies or exhibiting different sensitivities to task cues \citep{binz2023using}. Work on AI evaluation in expert and organizational settings likewise shows that output accuracy alone may fail to capture what practitioners need from AI-supported decisions in context \citep{wieringa2020means,lebovitz2021ai}. Moreover, work on chain-of-thought faithfulness shows that a model's stated reasoning can diverge from the features or intermediate reasoning that actually drive its output \citep{lanham2023measuring, turpin2023language}. Our empirical results extend these concerns to pluralistic alignment: in one domain, models with very different process-alignment scores achieve similar output accuracy, meaning that output-level evaluation can obscure which organizational values a model is actually implementing. Process-level measurement is therefore a necessary component of pluralistic alignment evaluation, particularly in high-stakes organizational deployment contexts.


\section{Method}


We use CALM to measure process alignment between an organizational benchmark and each LLM across three conditions.

\textbf{Baseline:} The model receives a structured case profile and decides without additional guidance. This reveals the model's \emph{implicit} cue-utilization policy, reflecting what it has learned through pretraining.

\textbf{Org-externalized:} The model receives the organization's cue-weighting policy, derived from ridge logistic regression on the organization's historical decisions, as explicit guidance before the case. High-weighted cues are described as strong indicators, medium-weighted cues as moderate, low-weighted cues as weak. This tests whether making the implicit past decision policies explicit can close the contextual knowledge gap and improve contextualized alignment.

\textbf{Introspective-externalized:} The model is shown how its own baseline cue-weighting policy diverges from the organization's (aggregate patterns across all cases) and instructed to self-correct. This tests whether self-directed adjustment can supplement or replace explicit guidance.

We estimate behavioral policies using ridge-regularized logistic regression on observed decisions, fitting identical cue sets to both the organizational benchmark and each LLM condition. Cue values are one-hot encoded and standardized to ensure coefficient comparability. The resulting coefficient vectors $\boldsymbol{\beta}$ represent each decision-maker's inferred cue-utilization policy. Alignment is measured as:
\begin{equation}
  \text{cos}(\theta) = \frac{\boldsymbol{\beta}_{\text{org}} \cdot \boldsymbol{\beta}_{\text{LLM}}}{\|\boldsymbol{\beta}_{\text{org}}\| \|\boldsymbol{\beta}_{\text{LLM}}\|}
\end{equation}

This metric is scale-invariant, symmetric, and bounded in $[-1, 1]$, where 1.0 indicates perfect alignment, 0 orthogonality, and negative values opposition. Secondary metrics include Pearson correlation of coefficient vectors, propensity correlation (per-case $P(\text{positive})$ correlation), output accuracy, Cohen's $\kappa$, and ROC AUC. We test significance via bootstrap permutation.

The procedure thus serves two functions depending on the domain. Where the organizational benchmark reflects a legitimate, well-specified normative policy, CALM functions as a calibration tool that quantifies policy divergence and guides externalization to close the gap. Where the benchmark encodes historically contested practices, meaning decision policies that conflict with contemporary legal or ethical standards, the same procedure functions as an audit tool. It renders the behavioral policy legible so that its normative implications can be examined. The two studies below illustrate both functions.


\section{Study 1: ECHR Article 6 Decisions}

\subsection{Context and Data}

We analyze 1,000 ECHR Article~6 decisions (50\% violation / 50\% no violation). Cues were coded as 45 binary features organized into nine families (Delay, Counsel, EvidenceAndArms, TribunalIntegrity, etc.) using GPT-5.4-mini with a codebook derived from Article~6 jurisprudence. We tested ten models selected to span sizes, providers, and geographical origin: GPT-5.4, GPT-5.4-mini, GPT-5.4-nano, Grok 4.1 Fast, Claude Haiku 4.5, Gemma 4 (31B), Minimax M2.7, Mistral Large, DeepSeek-v3.2, and Qwen 3.6 Plus, spanning US-, European-, and China-based providers. The overall predictive performance of the best models is consistent with prior computational studies of ECHR outcomes \citep{aletras2016predicting,chalkidis2021paragraph}. The court's ridge logistic regression achieves ROC AUC = 0.714, confirming that binary cues and linear model can capture meaningful but partial variance in judicial reasoning. 

\subsection{Baseline Process Alignment}

Baseline process alignment varies widely across the ten models. Cosine similarity ranges from $-0.211$ (GPT-5.4-nano) to $+0.844$ (GPT-5.4-mini). Three models cluster above $r_{\cos} = 0.82$: GPT-5.4-mini (0.844), Grok 4.1 Fast (0.842), and GPT-5.4 (0.824). Three others show moderate positive alignment: Qwen 3.6 Plus (0.750), Gemma 4 (31B) (0.710), and Minimax M2.7 (0.592). Four models fall near zero or below: Mistral Large (0.083), DeepSeek-v3.2 (0.062), Claude Haiku 4.5 ($-0.057$), and GPT-5.4-nano ($-0.211$). The process alignment on this dataset is highly correlated with accuracy (see Figure 1). The spread in alignment does not track general-purpose benchmark performance or pricing tier.

\subsection{Effect of Externalization}

Org-externalization moves the cosine estimate in the predicted direction for 8 of 10 models, producing statistically significant gains (one-sided $p < .05$) for three initially misaligned models: GPT-5.4-nano ($\Delta = +0.906$), Claude Haiku 4.5 ($\Delta = +0.682$), and Minimax M2.7 ($\Delta = +0.176$). Ceiling effects appear for models already well-aligned: GPT-5.4-mini ($\Delta = +0.045$, n.s.) and GPT-5.4 ($\Delta = +0.045$, n.s.), consistent with the hypothesis that externalization has little room to improve a model whose implicit policy already matches the organization's. Introspective externalization moves the point estimate in the predicted direction for 6 of 10 models; comparable gains appear for Claude Haiku 4.5 ($\Delta = +0.755$), but Grok 4.1 Fast regresses sharply ($\Delta = -0.346$, $p = .002$ in the worsening direction), its well-calibrated implicit policy disrupted by self-correction feedback.


\section{Study 2: German Credit Decisions}

\subsection{Context and the Fairness Tension}

The German Credit Dataset \citep{hofmann1994statlog} contains organizational decisions on credit ratings for consumer credits, with a 70\% Good / 30\% Bad base rate and 20 cues including numerical (loan duration, credit amount, age) and categorical (credit history, employment status, housing) attributes.

This domain introduces a tension absent from the ECHR study. Several cues are protected attributes under contemporary anti-discrimination law: \texttt{personal\_status\_sex}, \texttt{age\_years}, and \texttt{foreign\_worker}. Their use in decision-making is therefore normatively and legally sensitive, and may be impermissible or require special justification. However, as the organizational policy is derived from historical human decisions, it may encode discriminatory patterns. The original credit officers may have weighted these attributes in ways that disadvantaged women, older applicants, or foreign workers. High process alignment in this case is therefore not unconditionally desirable \citep{selbst2019fairness,dolata2022sociotechnical}. A model achieving $r_{\cos} \approx 1.0$ with the target policy may be faithfully reproducing institutional discrimination.

This surfaces a core pluralistic alignment question: whose values should the model align to? The organization's historically documented decision policy? Its stated ethical commitments? Contemporary regulatory standards? These targets may conflict, and CALM makes this conflict visible precisely because it measures the implicit \emph{actual} weighting policy rather than stated intent.

\subsection{Setup}

We apply five LLMs (Claude Haiku 4.5, GPT-5.4-mini, GPT-5.4-nano, Grok 4.1 Fast, DeepSeek-v3.2) under three conditions to all 1,000 cases. These five were selected to span the range of baseline alignment observed in Study 1, from strongly positive ($r_{\cos} > 0.5$) to negative, while limiting the computational cost of running all three conditions across 1,000 cases in this replication study; the remaining Study 1 models will be included in a full replication. To address the original dataset's class imbalance (70/30), we evaluate on a balanced subset of 600 cases (300 Good, 300 Bad; stratified random sample, seed = 42) for all comparisons. The logistic regression on the organization's decisions on this subset achieves 71.5\% accuracy and ROC AUC = 0.776, establishing the performance ceiling that can be reached through knowledge externalization interventions based on the linear cue utilization insights.

\begin{figure*}[t]
  \centering
  \includegraphics[width=\textwidth]{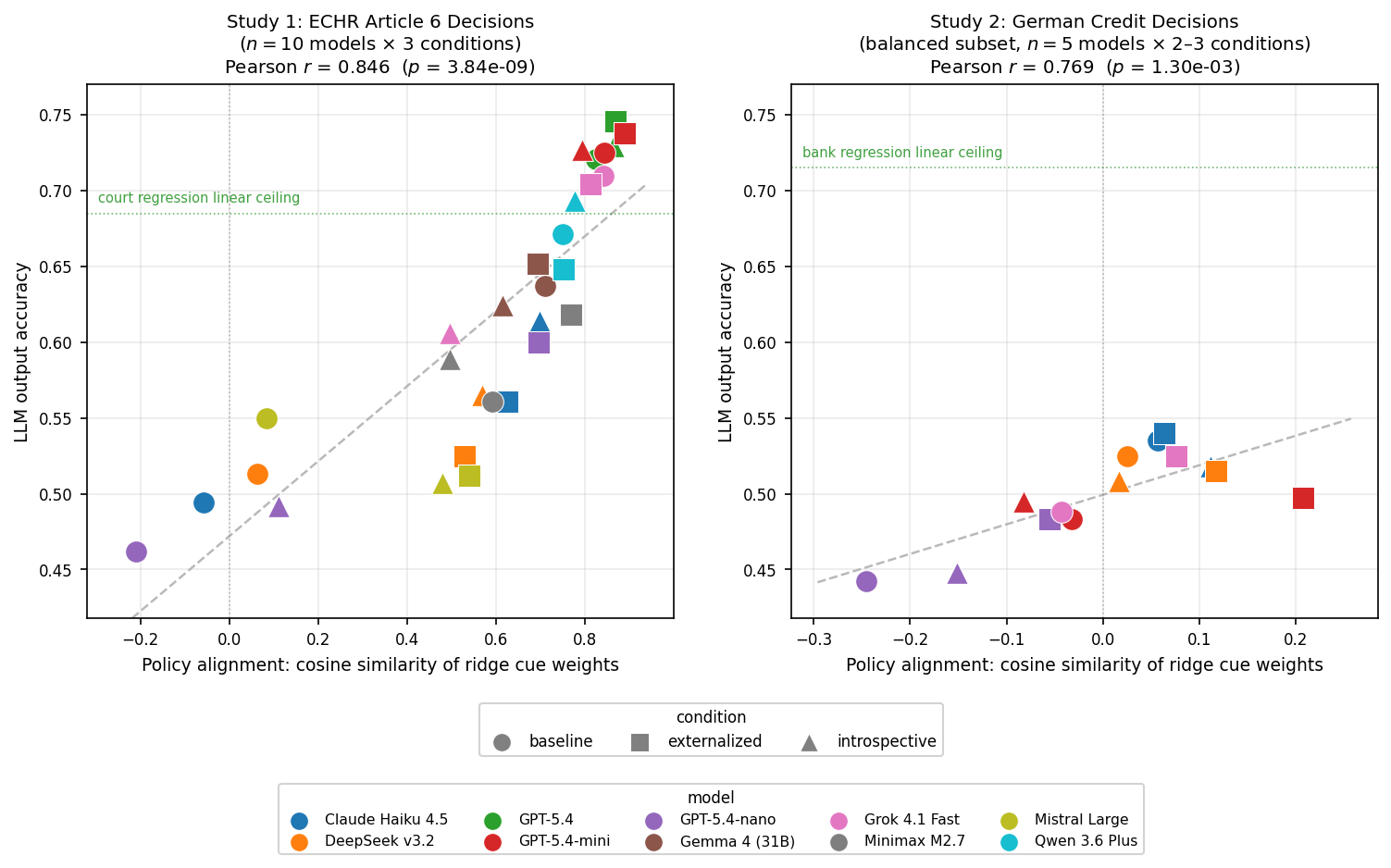}
  \caption{Policy alignment (cosine similarity) vs.\ output accuracy across both domains. \textbf{Left:} ECHR Article~6 decisions show a strong positive relationship ($r = 0.85$, $p < .001$, $n = 10$ models $\times$ 3 conditions). The dotted line marks the court regression linear ceiling. \textbf{Right:} German Credit decisions (balanced subset) show low but varied policy alignment overall (cosine $\in[-0.25,+0.21]$) with accuracy remaining near chance. The dotted line marks the bank regression linear ceiling.}
  \label{fig:scatter}
\end{figure*}

\subsection{Baseline Process Alignment}

Table~\ref{tab:german-baseline} reports baseline results. Cosine similarity is uniformly low (baseline range $-0.25$ to $+0.06$; $-0.25$ to $+0.21$ across all conditions), it still varies much more strongly than the output accuracy figure alone would suggest. This confirms the heterogeneous implicit policies of models already visible in the ECHR domain. However, as the process alignment is overall still very low, even for the best models, the positive Pearson correlation ($r=+0.77$, $p=.001$) is driven by a single model and falls to $+0.36$ when GPT-5.4-nano is excluded. This contrasts sharply with the ECHR domain ($r=0.85$), where the differences in process alignment reliably predict output accuracy.

\begin{table}[t]
  \caption{Baseline process alignment in the German Credit domain (balanced subset, $n=600$, corrected cue-value coefficients). Linear ceiling: 71.5\% accuracy, AUC = 0.776 (balanced subset, 5-fold CV). Good\% = proportion of cases classified as Good risk. AUC = predictability of the LLM's decisions from the cue values (5-fold CV).}
  \label{tab:german-baseline}
  \begin{center}
    \begin{small}
      \begin{tabular}{lcccc}
        \toprule
        Model & $r_{\cos}$ & Acc (\%) & AUC & Good\% \\
        \midrule
        Claude Haiku 4.5 & $+0.057$   & 53.5     & 0.907 &  9.2 \\
        GPT-5.4-mini     & $-0.033$   & 48.3     & 0.930 & 68.0 \\
        GPT-5.4-nano     & $-0.246$   & 44.2     & 0.944 & 50.5 \\
        Grok 4.1 Fast    & $-0.043$   & 48.8     & 0.939 & 37.5 \\
        DeepSeek-v3.2    & $+0.025$   & 52.5     & 0.930 &  5.5 \\
        \bottomrule
      \end{tabular}
    \end{small}
  \end{center}
\end{table}

The Good\% column reveals another anomaly: the implicit approval rates span 5.5\% (DeepSeek) to 68\% (GPT-5.4-mini), despite all models encountering the same balanced cases. DeepSeek and Claude Haiku classify almost all applicants as Bad risk; GPT-5.4-mini closely tracks the 70\% historical base rate. These are fundamentally different implicit policies about credit risk, yet they produce nearly identical accuracy figures. The cue utilizations made visible with CALM allow for deeper investigation of what drives these policies.

\subsection{Effect of Externalization}

Table~\ref{tab:german-extern} reports alignment changes under both interventions. Org-externalization shifts every model toward the organization's policy ($\Delta r_{\cos}$ from $+0.01$ to $+0.24$), but the shifts are small and leave alignment near zero (no cell exceeds $+0.21$). Introspective externalization is inconsistent (two of four models improve, two decline; $\Delta$ from $-0.05$ to $+0.09$) and negligible on average. Thus, even if the intervention seems to nudge the models in the right direction, it cannot lift their policies into substantive agreement with the benchmark and consequently the accuracy remains near chance.

\begin{table}[t]
  \caption{Change in process alignment under org-externalization and introspective-externalization ($\Delta r_{\cos}$, German Credit balanced subset). $^{\dag}$Grok introspective condition excluded: model classified 99.5\% of cases as Good (degenerate over-correction).}
  \label{tab:german-extern}
  \begin{center}
    \begin{small}
      \begin{tabular}{lcccc}
        \toprule
        & \multicolumn{2}{c}{Org-ext} & \multicolumn{2}{c}{Introspective} \\
        \cmidrule(lr){2-3}\cmidrule(lr){4-5}
        Model & $r_{\cos}$ & $\Delta$ & $r_{\cos}$ & $\Delta$ \\
        \midrule
        Claude Haiku 4.5 & $+0.063$ & $+0.007$ & $+0.111$ & $+0.054$ \\
        GPT-5.4-mini     & $+0.207$ & $+0.240$ & $-0.082$ & $-0.049$ \\
        GPT-5.4-nano     & $-0.056$ & $+0.191$ & $-0.152$ & $+0.094$ \\
        Grok 4.1 Fast    & $+0.076$ & $+0.119$ & —        & $^{\dag}$ \\
        DeepSeek-v3.2    & $+0.117$ & $+0.092$ & $+0.016$ & $-0.009$ \\
        \bottomrule
      \end{tabular}
    \end{small}
  \end{center}
\end{table}

\textbf{Grok's degenerate over-correction.} When given personalized deviation feedback noting its under-approval relative to the 70\% historical base rate, Grok 4.1 Fast classified 99.5\% of cases as Good risk in the introspective condition, starkly contrasting its prior 37.5\% approval rate. This over-correction renders cosine similarity analysis uninformative for this cell and illustrates a failure mode of introspective calibration: language models may struggle to respond to in-context calibration feedback, with some faring significantly worse than others.

\subsection{Protected Attribute Weighting}

The coefficient vectors let us measure protected-attribute reliance directly. We define each decision-maker's relative weight on an attribute as the share of its unit-normalized coefficient vector that falls on that attribute's columns. We focus on two protected attributes: personal-status-and-sex, and foreign-worker status. The organization's historical policy places non-negligible weight on both (personal-status-and-sex $0.16$, foreign-worker $0.13$), in the range of legitimate cues such as employment tenure ($0.17$), property ($0.16$), and housing ($0.16$). This confirms that the benchmark itself relies on protected attributes. At baseline the LLMs rely on both less than the organization does (means: personal-status-and-sex $0.08$, foreign-worker $0.11$).

Org-externalization transfers the legitimate risk cues but not the protected ones. It raises the LLMs' use of the legitimate cues toward the organization's policy: checking account rises from $0.18$ to $0.35$, loan duration from $0.06$ to $0.18$, and credit amount from $0.08$ to $0.16$. Protected-attribute usage does not follow. Foreign-worker usage falls for four of five models and rises for none (Claude $0.15$ to $0.03$, GPT-5.4-mini $0.12$ to $0.02$, GPT-5.4-nano $0.08$ to $0.03$, and DeepSeek $0.13$ to $0.10$), and stays at $0.08$ for Grok. Personal-status-and-sex usage remains flat (mean $0.08$ to $0.07$).

Models thus adopt the legitimate component of the organizational policy while resisting its reliance on protected attributes. This pattern is consistent with models holding domain knowledge that gender and nationality should not drive credit decisions, knowledge that competes with the organizational signal. It also illustrates the audit function of CALM: the intervention that imports the legitimate cues does not import the organization's protected-attribute weighting, which output-level metrics cannot surface.

We also examined the models' stated weighting, taken from the descriptive tier labels they assign each case. The stated picture differs from the behavioral one: Claude claims foreign-worker HIGH in $56.8\%$ of cases, whereas the fitted coefficients show the models under-using the protected attributes and reducing that use under externalization. Stated reasoning and behavioral policy therefore diverge on the protected attributes, a concrete instance of the faithfulness gap (Section~6.3).


\section{Discussion}

\subsection{Two Axes of Heterogeneity}

Our results reveal heterogeneity in organizational alignment along two axes. Models differ in baseline alignment within a single domain, with cosine similarity ranging from strongly positive to negative, and this spread tracks neither pricing nor general-purpose benchmark performance. Which model an organization deploys is therefore itself a choice among distinct implicit policies.

Organizations differ in the structure of alignment, and the cross-domain contrast is the clearest instance. In the ECHR domain, process alignment predicts output accuracy strongly ($r = 0.85$), models start from diverse baselines, and externalization reliably closes the gap for misaligned models. In the German Credit domain, the alignment-accuracy relationship does not hold reliably. Process alignment stays low overall but varies more than output accuracy, and the apparent positive correlation ($r = +0.77$). is driven by a single model, falling to +0.36 once GPT-5.4-nano is excluded. The models reach uniformly poor accuracy despite this variation in alignment (44--54\% against a 71.5\% ceiling), and interventions produce inconsistent results. This contrast is an important finding for steerable pluralism: the same measurement procedure, applied to two organizations, reveals qualitatively different alignment structures.

We argue this reflects a difference in the two organizational benchmarks. The ECHR's cue-weighting policy reflects accumulated legal doctrine, a relatively stable and publicly articulated standard that well-trained models can learn from pretraining and that externalization can reinforce. German consumer credit decisions in the 1990s combine legitimate risk signals (loan duration, credit amount, repayment burden) with historically contingent social judgments that anti-discrimination law has since partly superseded.

High process alignment with the German Credit policy is therefore not a clean target. A model that perfectly replicates the historical credit officer's policy may capture genuine risk signals, or it may reproduce discriminatory judgments about gender, age, and immigration status. CALM cannot adjudicate between these readings. Its contribution is to make the behavioral policy legible so that the question can be asked.

A natural objection is that the two domains were selected to produce this contrast, which would render the contestedness explanation post hoc. We address this directly. The domains were chosen on the basis of data availability, institutional prominence, and structural distinctiveness, before results were examined. The contestedness hypothesis carries empirical content. It predicts a strong alignment-accuracy relationship where the target policy reflects stable, publicly articulated standards, and a weak one where it reflects historically contingent social judgments. It would be falsified if a domain with clearly contested norms showed a strong alignment-accuracy correlation, or if a domain with a legitimate benchmark showed the orthogonality observed in German Credit. Testing CALM on clinical guidelines, hiring rubrics, and audit standards would let norm stability and discriminative ceiling vary independently, and would provide a more rigorous test of the mechanism.

\subsection{Steerable Pluralism Requires Process-Level Measurement}

Of the three modes of pluralistic alignment formalized by \citet{sorensen2024roadmap}, steerable pluralism is the most relevant to organizational deployment, because an organization deploying an LLM wants it to adopt their specified policy for a safe and predictable deployment. Our externalization conditions test this behaviorally, by asking whether a model's cue-weighting policy shifts toward a stated organizational policy when providing organizational knowledge to the LLMs via in-context learning. The answer is domain-dependent. In the legal domain, steering is partial but real, and it closes the gap for several misaligned models. In the credit domain, it largely fails.

Why steering fails in the credit domain is the central open question this work raises. One explanation is a knowledge gap: tacit organizational knowledge resists full articulation in a prompt \citep{polanyi1966tacit}, so externalization underspecifies the target policy. Two further explanations are harder to separate, because they share a behavioral signature while differing in normative valence. The models may resist the organizational policy because it encodes historical discrimination, in which case training-time safety alignment is working as intended. Alternatively, the models may be unable to adopt an arbitrary organizational policy at all, independent of its ethics, which would indicate a limit on steerable pluralism. We find indications of resistance in the credit domain, but we cannot yet attribute it to either cause. Distinguishing them, for example by testing whether resistance concentrates on protected attributes or appears uniformly across cues, is a priority for future work.

The credit task itself may be poorly suited to LLMs, which qualifies this interpretation. All five models perform near chance, at 44--54\% accuracy against a 71.5\% linear ceiling, so part of the observed failure to steer may reflect general task difficulty rather than a steerability limit specific to organizational alignment. Disentangling the two requires domains where models are individually competent yet still resist a stated organizational policy. Nevertheless, the model heterogeneity for implicit baseline policies and reactiveness to externalization interventions are confirmed in this replication and motivate further research.

Reliable steerability matters more in credit risk than in the legal domain, even though credit is where steering currently fails. Legal doctrine under Article~6 approximates a single publicly articulated standard, so between-organization variation in the target policy is small. However, credit-risk policy varies substantially across institutions, which differ in risk appetite, customer base, and regulatory regime. Between-organization plurality is therefore larger and more important in the credit domain.

We therefore suggest to consider (organizational) steerability as a design goal for foundation models. Foundation models should be steerable toward a specified and legitimate organizational policy. As future work, we propose an organizational steerability benchmark that pairs documented institutional policies with held-out decisions and scores how faithfully each model adopts a policy under externalization. A model that adopts a stated policy when the policy is legitimate, and resists when it is not, would be more useful and safer for organizational deployment than one that is uniformly inert or uniformly compliant.

\subsection{The Faithfulness Problem}

CALM infers cue-weighting policy from observed decisions, treating the model as a black box. It is worth emphasizing that this approach does not depend on chain-of-thought reliability. The estimated $\boldsymbol{\beta}_{\text{LLM}}$ captures the actual cue-weighting function the model implements across the full case set, derived from input-output pairs alone. This is the behavioral ground truth, independent of what the model claims to be doing in any individual reasoning trace. If anything, the potential divergence between stated reasoning and behavioral policy strengthens the case for CALM: a model can produce coherent, on-policy justifications while behaviorally weighting entirely different cues, meaning that auditing stated reasoning is insufficient and behavioral measurement is essential.

A separate but related question is whether a model's explicit chain-of-thought reasoning, produced when justifying a decision, reflects the same policy as its behavioral decisions. Prior work suggests systematic divergence: models can produce plausible-sounding reasoning that does not correspond to the features that actually drove their output \citep{lanham2023measuring, turpin2023language}. In the organizational alignment context, this creates a faithfulness problem with practical stakes. A model that explicitly cites the right cues in its reasoning while behaviorally weighting different ones would pass a surface-level audit but fail a process-alignment audit. Conversely, a model might weigh cues in alignment with the target policy while producing unfaithful explicit reasoning.

A natural extension of the present work is to compare CALM's behavioral policy estimates against cue weights extracted from explicit LLM reasoning, for instance by asking a scoring model to classify which cues were mentioned as influential in each case, then regressing those mention rates against the same cue set. If behavioral policy and stated reasoning track each other, then explicit reasoning is a reliable proxy for process alignment. If they diverge, auditing stated reasoning is insufficient and behavioral measurement via CALM is essential. We leave this analysis to future work.

\subsection{Process Alignment as an Audit Tool}

Our findings suggest a revised framing for CALM. In domains where legitimate, well-specified normative alignment targets are available (like ECHR), CALM is a \emph{calibration} tool: it measures how far a model deviates from the target policy and guides externalization to close the gap. In domains with contested normative benchmarks (like German Credit), CALM is primarily an \emph{audit} tool: it surfaces which protected and non-protected attributes each model is weighting, how those weights compare to the historical benchmark, and how they shift under intervention. This diagnostic use does not require a legitimate alignment target. Only the comparison needs to be informative.

For organizations deploying LLMs in high-stakes decision settings, CALM audits provide a third form of evaluation alongside output accuracy (are decisions correct?) and demographic disparity metrics (are decisions fair?): \emph{are decisions reached in the right way?} The importance of process legitimacy has long been recognized in procedural justice research \citep{tyler1990psychology}, and similarly has become important in consequential AI deployments \citep{selbst2019fairness,dolata2022sociotechnical}. CALM provides a scalable method for measuring one aspect of this process legitimacy from observed decisions.

\subsection{Limitations}

CALM estimates each decision policy as a linear cue-weighting function over a fixed set of coded cues, so its alignment scores are bounded by that representation. The cue set and the linear form are modeling choices, and both limit what the comparison can detect. Cues we do not code, and interactions a linear policy cannot express, are absent from the estimated weights and therefore from the cosine similarity. High alignment thus indicates agreement on the modeled cues, which is a proxy for equivalent decision processes rather than direct evidence of them. This estimate is faithful to observed decisions within the cue space, but a different cue design could shift the measured policy and the conclusions that follow from it.

The cross-domain contrast rests on two organizations, which limits what we can claim about its generality. Two domains are enough to show that alignment structure varies across organizations, but not to establish how common either pattern is. We cannot yet say whether the strong alignment-accuracy coupling in the legal domain or its absence in the credit domain is the more typical case. Testing CALM across domains that vary norm stability and discriminative ceiling independently, as we propose above, would show how the two patterns distribute rather than only that both occur.

\subsection{Policy Implications}

Our findings have direct relevance to emerging AI governance frameworks. The EU AI Act classifies consumer credit scoring as a high-risk AI application (Annex III), requiring conformity assessments, transparency obligations, and human oversight. Current regulatory guidance focuses primarily on output-level metrics: accuracy, demographic parity, and calibration. Our results suggest this is insufficient. Two systems with identical accuracy and disparity profiles can differ dramatically in \emph{how} they reach their decisions, with one replicating historically discriminatory weighting policies and the other not. CALM provides a process-level audit mechanism that can surface this divergence from observed decisions alone, without requiring access to model weights or internal representations.

More broadly, the question of \emph{whose} organizational policy an LLM should align with is itself a governance question. Organizations deploying AI must specify a normative target that has consequences beyond task performance only: which version of their decision policy should serve as the alignment reference? Should it be historical practice, stated ethical commitments, or current regulatory standards? CALM makes this choice explicit and auditable. We argue that pluralistic alignment frameworks should engage directly with these governance questions, since the technical problem of measuring alignment cannot be separated from the normative problem of deciding what to align to \citep{russell2019human,gabriel2020artificial}.


\section{Conclusion}

We compared process alignment across legal (ECHR) and financial (German Credit) decision domains using CALM as a behavioral method inspired by the Brunswik lens-model. The contribution for pluralistic alignment lies in the heterogeneity the comparison reveals. Across models, baseline alignment varies widely and tracks neither pricing nor benchmark tier. Across organizations, the alignment-accuracy relationship is strong and steerable where the benchmark reflects stable, publicly articulated norms, and unreliable where it reflects historically contingent and contested values. Both axes are difficult to detect with output-level evaluation, which motivates the use of CALM as one possible instantiation of the process-level measurement needed to identify them.

Organizational alignment thus is a pluralistic problem. It requires asking which organization, which policy, and from whose perspective. The same measurement procedure functions either as a calibration tool or as an audit tool, depending on the answer. Pluralistic alignment research should extend beyond between-person value diversity to include between-organization diversity, the partial articulability of implicit and tacit institutional knowledge, and the contestability of the norms that organizational alignment targets encode.


\section*{Impact Statement}

This work develops methods for measuring whether AI systems reason as specific organizations do. These methods have positive applications in auditing AI decision-making for process legitimacy and fairness. The same methods could, in principle, be used to align AI systems with organizations whose decision policies are discriminatory or harmful. We argue that surfacing this risk, as CALM does in the German Credit analysis, is preferable to leaving it invisible in output-level evaluations.


\bibliography{calm_references}
\bibliographystyle{icml2026}

\end{document}